\documentclass{article}

\usepackage{arxiv}

\usepackage[utf8]{inputenc} 
\usepackage[T1]{fontenc}    
\usepackage{hyperref}       
\usepackage{color,array}

\usepackage{graphicx}
\usepackage{multirow}
\usepackage{amsmath,amssymb,amsfonts}
\usepackage{mathtools}
\usepackage{adjustbox}
\usepackage{comment}

\graphicspath{{figures/}}

\title{Feature Construction Using Network Control Theory and Rank Encoding for Graph Machine Learning}

\author{
 Anwar Said \\
 Computer Science Department \\
 Vanderbilt University, TN, USA \\
 \texttt{anwar.said@vanderbilt.edu} \\
 \And
 Yifan Wei \\
 Computer Science Department \\
 Vanderbilt University, TN, USA \\
 \texttt{} \\
 \And
 Obaid Ullah Ahmad \\
 Electrical Engineering Department \\
 University of Texas at Dallas, Richardson, TX, USA \\
 \texttt{} \\
 \And
 Mudassir Shabbir \\
 Computer Science Department \\
 Information Technology University, Lahore, Pakistan \\
 \texttt{} \\
 \And
 Waseem Abbas \\
 Systems Engineering Department \\
 University of Texas at Dallas, Richardson, TX, USA \\
 \texttt{} \\
 \And
 Xenofon Koutsoukos \\
 Computer Science Department \\
 Vanderbilt University, TN, USA \\
 \texttt{} \\
}
\date{}
\begin{document}

\maketitle

\begin{abstract}
In this article, we utilize the concept of average controllability in graphs, along with a novel rank encoding method, to enhance the performance of Graph Neural Networks (GNNs) in social network classification tasks. GNNs have proven highly effective in various network-based learning applications and require some form of node features to function. However, their performance is heavily influenced by the expressiveness of these features. In social networks, node features are often unavailable due to privacy constraints or the absence of inherent attributes, making it challenging for GNNs to achieve optimal performance. To address this limitation, we propose two strategies for constructing expressive node features. First, we introduce average controllability along with other centrality metrics (denoted as NCT-EFA) as node-level metrics that capture critical aspects of network topology. Building on this, we develop a rank encoding method that transforms average controllability—or any other graph-theoretic metric—into a fixed-dimensional feature space, thereby improving feature representation. We conduct extensive numerical evaluations using six benchmark GNN models across four social network datasets to compare different node feature construction methods. Our results demonstrate that incorporating average controllability into the feature space significantly improves GNN performance. Moreover, the proposed rank encoding method outperforms traditional one-hot degree encoding, improving the ROC AUC from 68.7\% to 73.9\% using GraphSAGE on the GitHub Stargazers dataset,  underscoring its effectiveness in generating expressive and efficient node representations.
\end{abstract}


\section{Introduction}
\label{sec:introduction}

Network Control Theory (NCT) provides a rigorous mathematical framework for analyzing and influencing the dynamics of complex systems by leveraging network topology. It has been widely applied in domains such as biological networks, financial markets, and critical infrastructure systems \cite{gu2015controllability,friedland2012control}. At its core, NCT models inter-nodal communication as a diffusion process, where activity propagates over time \cite{pasqualetti2014controllability}. To regulate this diffusion, external control signals are introduced to steer the system toward a desired state while minimizing control energy. Understanding how network structure constrains or facilitates controllability offers valuable insights for optimizing system design and intervention. 

Graph Machine Learning (GML) builds on this foundation by providing a powerful approach to analyzing relational data, inherently capturing the complexities of dynamic systems \cite{hamilton2020graph}. Among GML techniques, Graph Neural Networks (GNNs) have gained prominence for learning on graph-structured data, using message passing to iteratively refine node representations for downstream tasks \cite{kipf2016semi}. This mechanism aggregates features from a node’s local neighborhood, encoding both local and global structural properties \cite{hamilton2017inductive}. However, the performance of GNNs is highly dependent on the quality and availability of node features. In many real-world settings, particularly in social networks, node features are sparse or entirely missing, limiting the expressiveness of the learned embeddings. A common workaround involves using one-hot encoding of node degrees as surrogate features \cite{xu2018powerful}. While this approach introduces basic structural signals, it often results in sparse, high-dimensional representations—especially in power-law networks—leading to increased computational complexity and suboptimal performance. Additionally, one-hot encoding is inherently restricted to discrete values and does not generalize well to continuous-valued structural metrics. These challenges highlight the need for a more expressive and computationally efficient feature initialization scheme.

To address this limitation, we propose leveraging \textit{average controllability}—a key metric in NCT—as a meaningful node feature. Average controllability quantifies a node’s ability to influence the system’s overall dynamics in response to external inputs, effectively characterizing its role in governing information flow across the network \cite{kalman1960contributions}. This metric offers valuable insight into how control energy propagates through the network and provides a principled basis for feature initialization. Unlike Laplacian positional encoding that rely on spectral information alone, average controllability captures reachability under linear dynamics and thus encodes causal influence. It enables the construction of node features that reflect not only topological structure but also a node’s potential to exert control over the network state. This may include a node’s ability to facilitate information diffusion or to reconfigure the network toward desired outcomes. The inclusion of such information is especially important for modeling tasks, as demonstrated in \cite{gasteiger2019diffusion}, where control-aware features significantly enhance learning performance.
Beyond average controllability, network science provides a broader suite of centrality measures that capture key structural characteristics of graphs. Closeness centrality identifies nodes that efficiently disseminate influence due to their proximity to others. Eigenvector centrality emphasizes nodes connected to highly influential neighbors, while betweenness centrality highlights bridge nodes that frequently mediate shortest paths, underscoring their role in controlling information flow \cite{freeman2002centrality}. These metrics encapsulate essential structural properties and have been widely adopted in modeling networked systems \cite{parkes2021network}.

By integrating average controllability with these centrality metrics, we introduce a principled approach to node feature construction that enhances the representational capacity of GNNs, particularly in scenarios where traditional node features are unavailable. Building on our previous conference work \cite{said2024improving}, this study systematically assesses the effectiveness of control-theoretic metrics in enhancing GNN performance and introduces a novel encoding scheme to integrate these metrics into learning pipelines. This work specifically extends the evaluation of the conference paper in three key aspects. First, it examines the impact of rank encoding by varying the number of bins, which is a critical parameter for features dimensionality. Second, it extends the evaluation of the proposed ranking-based encoding scheme to additional centrality metrics, including degree, closeness centrality, betweenness centrality, and eigenvector centrality along with the average controllability. Third, it expands the experimental setup by incorporating two additional social network datasets: Twitch Ego Net and Deezer Ego Net.

Our main contributions are as follows:


\begin{itemize}  
\item We incorporate graph-theoretic metrics, including average controllability, betweenness centrality, closeness centrality, and eigenvector centrality, as node features within the GNN framework.  

\item We propose a rank encoding method to map graph metrics, such as average controllability, into the node feature space. This approach enables the seamless transformation of any graph or node metric into a feature representation that can be directly utilized by GNNs for training. 

\item In the context of social network classification, where node features are often unavailable, we evaluate six GNN models using our proposed approaches. Through extensive experiments, we demonstrate the effectiveness of the proposed scheme in enhancing classification performance.  
\end{itemize}  

By incorporating average controllability alongside additional graph metrics and introducing our novel rank encoding scheme, this study aims to bridge the gap between NCT and GNNs, enhancing their predictive capabilities on unattributed networks. 

The structure of the manuscript is outlined below. Section \ref{sec:related-work} offers a brief review of the literature on NCT, GNNs and their integration. Section \ref{sec:method} presents our proposed methodology in depth. This is followed by Section \ref{sec:evaluation} which evaluate the proposed approaches put forth through various experiments and evaluation. The paper concludes with Section \ref{sec:conclusion} suggesting promising avenues for future research.

\section{Related Work}
\label{sec:related-work}

The integration of NCT with graph machine learning is an emerging research direction that holds significant potential for enhancing learning algorithms by incorporating dynamical and structural insights from complex networks.
In \cite{said2023network}, we employed a range of controllability metrics derived from the controllability Gramian to construct expressive representations of graphs.
This prior work considers metrics such as the trace, rank, and the first and last three eigenvalues of the controllability Gramian as potential features for graph representations. It also explores the use of varying numbers of control nodes (leaders), repeating the process to derive a diverse and expressive set of features for the final graph representation.
More recently, the authors of \cite{parkes2023using} introduced a Python package implementing several NCT metrics. In the context of brain networks, they investigated how connectome topology influences neural dynamics, evaluated the consistency of NCT outputs with empirical data on brain function and stimulation, and examined how these metrics vary across developmental stages and correlate with behavioral and mental health indicators. A broader application of NCT metrics to brain networks is also detailed in their earlier work \cite{kim2020linear}, which provides comprehensive methodological and empirical insights. 

In conjunction with NCT, the authors in \cite{yang2024control} introduce a novel paradigm for enhancing the performance of GNNs. They conceptualize the GNN as a control system, treating node features as controlled variables and predictions as the system state. By designing controllers that optimize these variables, they achieve improved predictive performance. Furthermore, they develop a neural controller tailored specifically for enhancing GNNs and introduce a neural Lyapunov function to ensure controller stability, enabling reliable adjustments of node features during training.

In GML, two main approaches are graph embedding methods and GNNs.
Graph embedding techniques rely on graph-theoretic and statistical tools to derive fixed-size representations of graphs or their components, which are subsequently used with traditional machine learning models such as support vector machines or random forests \cite{kriege2020survey}. Notable recent methods in this direction include Wasserstein-based and multiscale embeddings \cite{togninalli2019wasserstein, kondor2016multiscale}.
In contrast, GNNs are deep learning models trained in an end-to-end fashion \cite{kipf2016semi}. They leverage message passing, spectral techniques, and recurrent neural networks to learn node and graph-level representations \cite{hamilton2017inductive, defferrard2016convolutional, bruna2013spectral, bresson2017residual}. 
Building on these foundational ideas, several influential GNN architectures have emerged. Graph Attention Networks (GATs) apply attention mechanisms to dynamically weigh the importance of neighboring nodes, thereby enhancing representational expressiveness \cite{velivckovic2017graph}. GraphSAGE introduces a neighborhood sampling strategy that facilitates inductive learning on large-scale graphs \cite{hamilton2017inductive}. The work in \cite{morris2019weisfeiler} unifies various convolutional GNN models under a general framework, offering theoretical insights into their expressiveness. Other noteworthy advancements include transformer-based GNNs \cite{yun2019graph}, as well as graph pooling architectures such as DiffPool \cite{ying2018hierarchical} and EdgePool \cite{diehl2019edge}, which enable hierarchical graph representation learning.


In network settings where node features are absent, several strategies have been proposed to improve GNN performance \cite{wu2020comprehensive}. For example, the Local Degree Profile incorporates degree information into node feature vectors, thereby embedding additional structural context \cite{cai2018simple}. Similarly, one-hot degree encoding represents node degrees as sparse binary vectors, offering a straightforward yet limited feature representation \cite{xu2018powerful}. Other approaches include centrality-based encoding, which reflect a node's positional significance within the graph, and motif-based features, which capture local subgraph patterns \cite{lee2019attention}. These techniques aim to inject topological information into the node feature space to enhance learning. However, to the best of our knowledge, the incorporation of controllability-based information into node features has not yet been explored in conjunction with GNNs—this forms the central focus of our study.

In an unattributed network setting, a number of techniques have been developed to encode the structural roles or positions of nodes in graphs that lack informative attributes. Spectral methods based on the eigenvectors of the graph Laplacian are commonly used to define positional embeddings, offering a global view of the graph's topology through its frequency modes~\cite{belkin2003laplacian, dwivedi2023benchmarking}. These encodings are particularly useful in transformer-style graph models where absolute or relative node positions help guide attention mechanisms. In contrast, structural role embeddings such as \textit{Struc2Vec}~\cite{ribeiro2017struc2vec} and \textit{Role2Vec}~\cite{ahmed2019role2vec} aim to learn representations that reflect the structural similarity of nodes, irrespective of their location in the graph. Other methods like \textit{GraphWave}~\cite{donnat2018learning} model the diffusion of signals across the graph to capture functional roles of nodes. Recent work such as \textit{LOTR}~\cite{ma2021unified} provides a unified framework that connects position- and role-based representations through learned node similarities. 

Building upon recent efforts such as \cite{yang2024control, said2023network}, we leverage NCT metrics to enhance GNN performance in unattributed graph settings. In particular, we investigate the utility of average controllability—a metric that quantifies a node’s ability to influence global system dynamics—in constructing informative node features. Additionally, we introduce a novel rank encoding scheme for node feature initialization. This method employs a histogram-based representation and reverse encoding, enabling the transformation of any scalar graph metric into a fixed-dimensional, expressive node feature vector.
   
\section{Methodology}
\label{sec:method} 

In this work, we incorporate NCT metrics—particularly average controllability—alongside other graph-theoretic measures to enrich the graph structure via node features for social networks. Starting from defining basic notations, the following sections describe the details of the proposed approaches. 


\subsection{Preliminaries}
\label{subsec:prelim}

Let \( G = (V, E, X) \) be an undirected graph representing a social network, where \( V \) is the set of $n$ users, \( E \subseteq V \times V \) represents the interactions between them, and \( X \in \mathbb{R}^{n \times d} \) is the node feature matrix, with each node having a feature vector of dimension \( d \). An interaction between two users \( u \) and \( v \) is denoted by the unordered pair \( (u, v) \). The \textit{neighborhood} of a user \( u \) is defined as \( \mathcal{N}_{u} = \{ v \in V \mid (u, v) \in E \} \), and the corresponding \textit{degree} is \( \deg(u) = |\mathcal{N}_{u}| \). The learned representation for a node \( v \) is denoted by \( \mathbf{h}_v \), while \( \mathbf{h}_G \) represents the learned representation of the entire graph.

A graph with \( n \) nodes can be represented by an adjacency matrix \( \mathbf{A} \in \{0,1\}^{n \times n} \), where \( \mathbf{A}_{i,j} = 1 \) if and only if \( (v_j, v_i) \in E \), and \( \mathbf{A}_{i,j} = 0 \) otherwise. The identity matrix is denoted by \( \mathbf{I} \). The matrix \( \mathbf{B} \in \{0,1\}^{n \times m} \) encodes how control inputs affect the nodes, where \( \mathbf{B}_{i,j} = 1 \) indicates that the \( j \)-th control input is applied to node \( i \). Additionally, \( \mathbf{u}(t) \in \mathbb{R}^{m\times 1} \) denotes an input vector, \( \mathcal{W} \) represents the controllability Gramian matrix, and \( \mathcal{C}(\mathbf{A}, \mathbf{B}) \in \mathbb{R}^{n\times 1}\) is a vector containing the average controllability values for all nodes in graph \( G \) for the control input \(\mathbf{B}\).

\vspace{0.1in}
\textbf{Graph Classification:}\\

Graph classification is a fundamental task in graph machine learning. It involves assigning a categorical label to an entire graph, indicating its membership in a specific class. This task has numerous real-world applications, including cheminformatics, where molecules are classified based on their structure; document categorization, where documents or articles are grouped based on their content and relationships; and social network analysis, where patterns of social interactions are examined \cite{hamilton2017inductive}.

Formally, consider a collection of graphs \( \mathcal{G} = \{G_1, G_2, \ldots, G_k\} \), where each graph \( G_i = (V_i, E_i, X_i) \) consists of a node set \( V_i \), edge set \( E_i \), and node feature matrix \( X_i\). Each graph \( G_i \) is associated with a label \( y_i \in \mathcal{Y} = \{0, 1, \ldots, C\} \), where \( C \in \mathbb{N} \) denotes the number of unique classes. The objective is to learn a classification function \( \phi: \mathcal{G} \rightarrow \mathcal{Y} \), which assigns a predicted label \( \tilde{y}_i = \phi(G_i) \) to each input graph.

The machine learning approach to graph classification involves learning the parameters \( \boldsymbol{\theta} \) of a model that approximates the classification function \( \phi \). The model \( \phi(G, \boldsymbol{\theta}) \) is trained to minimize a loss function \( \mathcal{L}(y, \tilde{y}) \), which quantifies the discrepancy between the predicted label \( \tilde{y} \) and the true label \( y \), thereby guiding the model toward improved classification accuracy.


Among graph machine learning techniques, GNNs have emerged as the dominant approach for learning from graph-structured data. GNNs use a message passing mechanism to iteratively refine node representations by aggregating information from their local neighborhoods \cite{kipf2016semi}. This process allows GNNs to capture both local and global structural patterns in the graph \cite{hamilton2017inductive}. However, the effectiveness of this approach is heavily reliant on the availability and quality of the input node feature matrix \( X \), which serves as the input to the message passing process.

We discuss the architecture and working of GNNs in more detail in Section~\ref{subsec:gnns}. Here, we emphasize a critical challenge: in many real-world graphs, node features may be incomplete or entirely unavailable. This can result from privacy constraints, the absence of intrinsic attributes, the high cost of data acquisition, or the nature of the data itself, where nodes lack meaningful descriptors. 

In such settings, a key question arises: \emph{how can we train GNNs effectively when node features are missing?} We posit that enriching the graph with expressive initial node features can significantly improve GNN performance. To this end, we propose using average controllability along with other graph centralities and rank encoding to construct informative node representations. In the following sections, we detail our methodology for generating such features using the proposed approaches.


\begin{figure*}[!t]
    \centering
    \includegraphics[width=\textwidth]{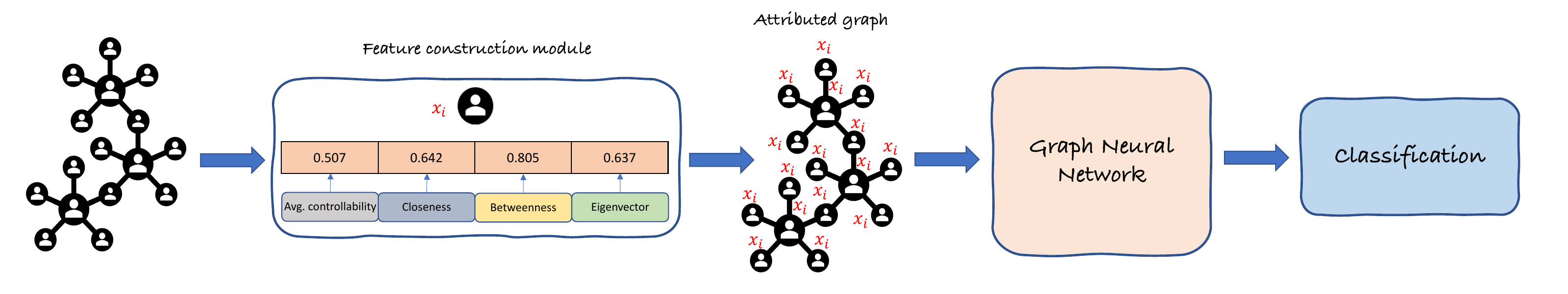}
    \caption{Illustration of the proposed NCT-EFA framework: Starting with a simple graph, NCT-EFA leverages network control theory and network science metrics to enrich node features. These enhanced graphs are then used to train a GNN for the downstream classification task. Here, \( x_i \in X \) refers to the node feature initialized in the feature construction module.
}
    \label{fig:architecture}
\end{figure*}

\subsection{Constructing Node Features with Network Control Theory}
\label{subsec:node-features-construction}

In network settings where no node features are available, we propose two methods discussed below to construct expressive node features for GNNs.   

\subsubsection{Network Controllability Metrics}
\label{subsec:nctmetrics}

NCT allows us to identify influential nodes within a network—those capable of significantly shaping system dynamics. In the context of social networks, such nodes are analogous to individuals who can effectively disseminate information or influence collective behavior. By incorporating average controllability into node representations, we aim to encode not only topological information but also a node’s ability to steer the network toward desired states. This ability may involve strategically diffusing information or reconfiguring network interactions. 

NCT models a dynamic network system using an adjacency matrix \( \mathbf{A} \) and a control set \( \mathbf{B} \). Within this framework, the temporal evolution of node states is captured by a linear time-invariant system described by the differential equation:

\begin{equation}
    \label{eq:control}
    \dot{x}(t) = \mathbf{A}\boldsymbol{x}(t) + \mathbf{B}\boldsymbol{u}(t),
\end{equation}

where \( \boldsymbol{x}(t) = [x_1(t), x_2(t),\ldots, x_n(t)]^\top \) represents the vector of node states, \( \mathbf{A} \in \mathbb{R}^{N \times N} \) is the adjacency matrix of the graph, \( \boldsymbol{u}(t) \in \mathbb{R}^m \) is the vector of control inputs, and \( \mathbf{B} \in \mathbb{R}^{N \times m} \) defines how inputs are applied to the nodes. In our experiments, we set \( \mathbf{B} = \mathbf{I} \), the identity matrix, indicating full control over all nodes.

To quantify how easily a system can be driven from one state to another using control inputs, we compute the controllability Gramian \( \mathcal{W} \), a fundamental matrix in control theory \cite{pasqualetti2014controllability,ahmad2024control}. For the system in Equation~\ref{eq:control}, the infinite-horizon controllability Gramian is defined as:

\begin{equation}
    \label{eq:Gramian}
    \mathcal{W}(\mathbf{A}, \mathbf{B}) = \int_{0}^{\infty} e^{-\mathbf{A}\tau} \mathbf{B}\mathbf{B}^\top e^{-\mathbf{A}^\top \tau} \, d\tau \in \mathbb{R}^{N \times N}.
\end{equation}

If the system is stable (i.e., all eigenvalues of \( \mathbf{A} \) have negative real parts), the integral converges, and the Gramian can also be computed by solving the Lyapunov equation:

\begin{equation}
    \label{eq:Lyapunov}
    \mathbf{A}\mathcal{W} + \mathcal{W}\mathbf{A}^\top + \mathbf{B}\mathbf{B}^\top = 0.
\end{equation}

This formulation provides a foundation for deriving various NCT metrics. Among them, \emph{average controllability} is of central interest in this study.

\vspace{0.05in}
\noindent\textbf{Average Controllability:} Average controllability quantifies a node’s ability to influence the network state in response to control inputs \cite{parkes2021network}. A node with high average controllability can effectively distribute control energy across the network, thereby exerting a greater influence on global dynamics. It is computed as the diagonal entries of the controllability Gramian:

\begin{equation*}
    \label{eq:avg-cont}
    \mathcal{C}(\mathbf{A}, \mathbf{B}) = \mathrm{diag}(\mathcal{W}(\mathbf{A}, \mathbf{B})),
\end{equation*}

where \( \mathcal{C} \in \mathbb{R}^{N \times 1} \) is a vector of average controllability values, and \( \mathcal{C}(v) \) represents the controllability of node \( v \). Because this metric captures a node’s potential impact on network dynamics, we hypothesize that incorporating it into node features can improve the quality of embeddings learned by GNNs.


In this study, we compute the average controllability of network nodes using a finite-horizon formulation of the controllability Gramian. Rather than relying on the infinite-horizon solution of the continuous-time Lyapunov equation-which requires the system matrix \( A \) to be Hurwitz (i.e., all eigenvalues with negative real parts) to ensure convergence—we adopt a numerical integration approach over a bounded time interval \([0, T=1, step=0.001]\). This method enables us to handle undirected, symmetric, and potentially unstable adjacency matrices commonly observed in real-world networks such as social graphs. Specifically, we compute the finite-horizon Gramian by simulating the system dynamics over discrete time steps and integrating the resulting matrix sequence. This formulation is well-defined regardless of the spectral properties of \( A \), and thus accommodates networks that may be marginally stable or even unstable. Additionally, we use a full control input matrix \( B = I \), which assumes actuation at all nodes. 

In the following, we describe four additional graph-theoretic metrics used to further enrich node features.


\noindent\textbf{One-Hot Degree Encoding:}  
One-hot degree encoding is a method for representing node degrees as categorical features in a binary format. Each node's degree is encoded as an \( n \)-dimensional sparse vector, where only one entry corresponding to the degree value is set to 1, and all other entries are 0. This approach has been widely used in the GNN literature \cite{xu2018powerful}. However, its drawbacks include high sparsity and increased dimensionality as the range of degrees grows, which can adversely impact scalability and learning. We use this method as a baseline in our experiments.

\vspace{0.1cm}

\noindent\textbf{Closeness Centrality:}  
Closeness centrality measures how close a node is to all other nodes in the graph. It is defined as the reciprocal of the average shortest path distances from a given node to all others. Formally, for a node \( v \in V \), the closeness centrality is given by:
\begin{equation}
    C_{\text{closeness}}(v) = \frac{n-1}{\sum_{u} d(u, v)},
\end{equation}
where \( d(v, u) \) is the length of the shortest path between nodes \( v \) and \( u \). Nodes with high closeness centrality can efficiently interact with all others and are typically central to the network's communication structure \cite{wasserman1994social}.

\vspace{0.1cm}

\noindent\textbf{Betweenness Centrality:}  
Betweenness centrality quantifies how often a node appears on the shortest paths between other pairs of nodes. It reflects the node’s ability to control information flow within the network. For a node \( v \), it is defined as:
\begin{equation}
    C_{\text{betweenness}}(v) = \sum_{s \neq v \neq t \in V} \frac{\sigma_{st}(v)}{\sigma_{st}},
\end{equation}
where \( \sigma_{st} \) is the total number of shortest paths from node \( s \) to node \( t \), and \( \sigma_{st}(v) \) is the number of those paths that pass through \( v \). Nodes with high betweenness centrality often act as bridges and can significantly influence the flow of information across the network \cite{wasserman1994social}.

\vspace{0.1cm}

\noindent\textbf{Eigenvector Centrality:}  
Eigenvector centrality expands upon the degree centrality and measures a node’s influence on both the importance of the node itself and the quality of its neighbors. It assigns higher scores to nodes connected to other important nodes in the graph. Let \( \mathbf{A} \) be the adjacency matrix of the graph and $C$ to be the eigenvector centrality vector. With small rearrangement, the eigenvector centrality can be rewritten in the form of eigenvector equation:
\begin{equation}
    \mathbf{A}C = \lambda C,
\end{equation}


Building upon the four structural metrics—average controllability, closeness centrality, betweenness centrality, and eigenvector centrality—we compute each metric for every node in the graph and concatenate them into a composite feature vector. This vector serves as a compact yet expressive representation of a node's structural role, capturing aspects such as influence, reachability, diffusion and control potential. These node-level features are then used as inputs for training the GNN.

Figure~\ref{fig:architecture} provides an illustrative overview of how these features are computed, encoded, and assembled into the final node representations for the graph classification task.

\begin{figure*}
    \centering
    \includegraphics[width=1.0\linewidth]{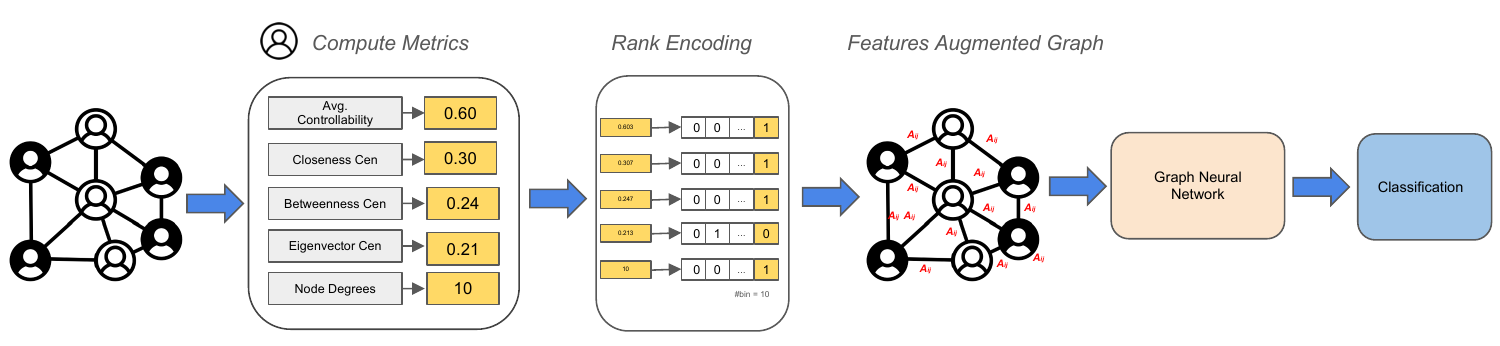}
    \caption{Illustration of the proposed rank encoding scheme. Initially, desired centrality metrics are computed. Each metric is then transformed into a one-hot encoded vector. These encoding are concatenated to form a unified feature vector for each node. The augmented graph is subsequently used to train a GNN for the social network classification task.}
    \label{fig:rank-encoding}
\end{figure*}

\subsubsection{Rank Encoding}
\label{subsec:rankEnc}


As discussed earlier, constructing expressive node features is crucial in settings where node attributes are unavailable. Existing encoding strategies for unattributed graphs primarily fall into two categories. The first, and most widely used, is one-hot degree encoding, where each node’s degree is encoded as a sparse binary vector \cite{xu2018powerful}. The second approach involves directly concatenating multiple centrality measures into a single feature vector \cite{cai2018simple}. While these techniques inject structural information, they either suffer from high dimensionality and sparsity or fail to capture the relative distribution of scalar node-level metrics across the graph.

To address these limitations, we introduce a novel \textit{rank-based encoding scheme} that converts any graph-theoretic metric—such as average controllability—into a structured, fixed-dimensional node feature vector. Our approach aims to preserve the relative distribution of the metric values across the graph while offering greater control over feature sparsity and dimensionality.
The proposed method combines a histogram-based encoding strategy with reverse rank mapping. Specifically, we first compute a graph-level histogram from the given centrality vector and then map each node’s scalar metric value to its corresponding histogram bin. The resulting bin index is used to generate a one-hot encoded feature vector for that node. Two input parameters—\( k \), the number of histogram bins, and the value range—govern the resolution and sparsity of the encoding.

For example, given the average controllability vector \( \mathcal{C} \) from Section~\ref{subsec:nctmetrics}, where each element \( \mathcal{C}(v) \) denotes the controllability of node \( v \), we construct a histogram \( \mathcal{H} \) with \( k \) bins covering the full range of controllability values. Each bin represents a subrange, and its height reflects the number of nodes with metric values in that range. To derive a node-level feature vector \( \mathbf{h}^0_v \), we identify the bin \( \mathcal{H}(i) \) into which \( \mathcal{C}(v) \) falls and assign a one-hot vector accordingly:

\begin{equation*}
\small
\mathbf{h}^0_v(i) = 
\begin{cases} 
1 & \text{if } \mathcal{C}(v) \in \mathcal{H}(i), \\
0 & \text{otherwise}.
\end{cases}
\end{equation*}


To illustrate the process of rank encoding, consider a simple example with a graph consisting of five nodes, each with an associated average controllability value from the vector \( \mathcal{C} \): \( [0.1, 0.3, 0.5, 0.7, 0.9] \). Suppose we construct a histogram \( \mathcal{H} \) with three bins to cover these values: Bin 1 for the range [0.0, 0.3), Bin 2 for [0.3, 0.6), and Bin 3 for [0.6, 1.0]. Node 1, with a controllability value of 0.1, falls into Bin 1, resulting in a one-hot vector of [1, 0, 0]. Node 2, with a value of 0.3, also maps to Bin 1 and thus shares the same one-hot vector. Node 3, with a value of 0.5, falls into Bin 2, giving it a vector of [0, 1, 0]. Nodes 4 and 5, with values 0.7 and 0.9, respectively, are placed in Bin 3, resulting in vectors of [0, 0, 1]. 

Figure~\ref{fig:rank-encoding} illustrates the complete encoding pipeline. The rank encoding module first computes histograms for each centrality metric. These histograms are then used to transform node-level metric values into one-hot encoded feature vectors using the reverse mapping strategy. When multiple metrics are used, each is independently encoded and then concatenated to form a comprehensive node representation.

This method is inherently \textit{type-agnostic}: it can encode any scalar node attribute, including continuous metrics like average controllability or discrete ones like node degree, without requiring the feature dimension \( d \) to match the maximum value in the dataset. The resulting augmented graph—with expressive, structured node features—is subsequently used as input to the GNN for training.

\begin{table*}[!t]
\centering
\small
\caption{Dataset statistics}
\begin{tabular}{|l|c|cccc|cc|cc|c|c|}
\hline
\multirow{2}{*}{\textbf{Dataset}} & \multirow{2}{*}{\textbf{Graphs}} & \multicolumn{4}{c|}{\textbf{Nodes}}          & \multicolumn{2}{c|}{\textbf{Density}}       & \multicolumn{2}{c|}{\textbf{Diameter}} & \textbf{Classes} & \textbf{Task} \\ \cline{3-10}
                                  &                                  & \multicolumn{1}{c|}{Min} & \multicolumn{1}{c|}{Max} & \multicolumn{1}{c|}{Mean} & Median & \multicolumn{1}{c|}{Min}   & Max   & \multicolumn{1}{c|}{Min} & Max & & \\ \hline
Reddit Threads                    & 203,088                          & \multicolumn{1}{c|}{11}  & \multicolumn{1}{c|}{97}   & \multicolumn{1}{c|}{23.93} & 17    & \multicolumn{1}{c|}{0.021} & 0.382 & \multicolumn{1}{c|}{2}   & 27  & 2 & Graph Classification \\ \hline
GitHub Stargazers                & 12,725                           & \multicolumn{1}{c|}{10}  & \multicolumn{1}{c|}{957}  & \multicolumn{1}{c|}{113.79} & 47    & \multicolumn{1}{c|}{0.003} & 0.561 & \multicolumn{1}{c|}{2}   & 18  & 2 & Graph Classification \\ \hline
Twitch Ego Nets                  & 127,094                          & \multicolumn{1}{c|}{14}  & \multicolumn{1}{c|}{52}   & \multicolumn{1}{c|}{29.67} & 28    & \multicolumn{1}{c|}{0.038} & 0.967 & \multicolumn{1}{c|}{1}   & 2   & 2 & Graph Classification \\ \hline
Deezer Ego Nets                  & 9,629                            & \multicolumn{1}{c|}{11}  & \multicolumn{1}{c|}{363}  & \multicolumn{1}{c|}{23.19} & 17    & \multicolumn{1}{c|}{0.015} & 0.909 & \multicolumn{1}{c|}{2}   & 2   & 2 & Graph Classification \\ \hline
\end{tabular}
\label{tab:dataset-stats}
\end{table*}

\begin{table*}[!htb]
\centering
\caption{ROC AUC Comparison (Mean ± Std) of degree Encoding vs. NCT-EFA. \textbf{Bold} represent the methods with best performance.}
\label{tab:results_deg_nct}
\resizebox{\textwidth}{!}{%
\begin{tabular}{|l|cc|cc|cc|cc|cc|cc|}
\hline
\multirow{2}{*}{Datasets} 
& \multicolumn{2}{c|}{\textbf{GraphConv}} 
& \multicolumn{2}{c|}{\textbf{GraphSAGE}} 
& \multicolumn{2}{c|}{\textbf{GCN}} 
& \multicolumn{2}{c|}{\textbf{UniMP}} 
& \multicolumn{2}{c|}{\textbf{ResGatedGCN}} 
& \multicolumn{2}{c|}{\textbf{GAT}} \\
\cline{2-13}
& \textit{deg} & NCT-EFA & \textit{deg} & NCT-EFA & \textit{deg} & NCT-EFA & \textit{deg} & NCT-EFA & \textit{deg} & NCT-EFA & \textit{deg} & NCT-EFA \\
\hline

Reddit Threads        
& 83.14 ± 0.11 & \textbf{84.04 ± 0.06} 
& 82.86 ± 0.24 & \textbf{83.77 ± 0.04} 
& 82.74 ± 0.22 & \textbf{83.48 ± 0.05} 
& 83.55 ± 0.05 & \textbf{83.74 ± 0.03} 
& 83.68 ± 0.08 & \textbf{84.07 ± 0.05} 
& 83.44 ± 0.03 & \textbf{83.59 ± 0.04} \\ 

GitHub Stargazers     
& 71.47 ± 0.56 & \textbf{76.14 ± 0.62} 
& 68.70 ± 1.40 & \textbf{73.95 ± 0.30} 
& 70.53 ± 0.52 & \textbf{71.34 ± 0.60} 
& 68.75 ± 0.65 & \textbf{74.68 ± 0.35} 
& 72.63 ± 0.53 & \textbf{76.57 ± 0.46} 
& 67.30 ± 1.82 & \textbf{72.83 ± 0.30} \\ 

Deezer Ego Net      
& \textbf{53.93 ± 1.26} & 51.13 ± 1.42 
& \textbf{53.29 ± 1.20} & 51.68 ± 2.22 
& 51.31 ± 2.20 & \textbf{53.18 ± 1.23} 
& \textbf{54.55 ± 0.88} & 51.08 ± 0.93 
& \textbf{54.30 ± 1.55} & 51.47 ± 3.09 
& \textbf{53.30 ± 1.57} & 52.70 ± 0.99 \\ 

Twitch Ego Net      
& 65.26 ± 0.05 & \textbf{65.70 ± 0.04} 
& 64.68 ± 0.05 & \textbf{65.43 ± 0.06} 
& 64.71 ± 0.10 & \textbf{65.27 ± 0.09} 
& 64.76 ± 0.09 & \textbf{65.54 ± 0.04} 
& 65.28 ± 0.12 & \textbf{65.68 ± 0.04} 
& 65.16 ± 0.05 & \textbf{65.27 ± 0.04} \\ 
\hline
\end{tabular}%
}
\end{table*}

\subsection{Graph Neural Networks}
\label{subsec:gnns}

Graph Neural Networks (GNNs) are a class of neural architectures designed to operate on graph-structured data \cite{hamilton2020graph}. They work by aggregating information from a node’s local neighborhood, allowing the network to learn both local and global structural dependencies \cite{kipf2016semi}. A widely adopted framework that generalizes many GNN variants is the Message Passing Neural Network (MPNN). In this framework, node representations are updated over multiple layers by aggregating features from neighboring nodes. This process enables each node to learn a contextualized embedding that reflects its position and role within the graph. The message passing procedure is typically defined as:

\begin{equation*}
\small
    \boldsymbol{h}_v^{(t+1)} = \text{UPDATE} \left( \boldsymbol{h}_v^{(t)}, 
    \text{AGGREGATE} \left( \{ \boldsymbol{h}_u^{(t)} : u \in \mathcal{N}(v) \} \right) \right),
\end{equation*}
where \( \boldsymbol{h}_v^{(t)} \) denotes the feature vector of node \( v \) at iteration \( t \), \( \mathcal{N}(v) \) is the set of neighbors of \( v \), and \texttt{AGGREGATE} and \texttt{UPDATE} are learnable or predefined functions that control how neighborhood information is combined and applied.

The strength of MPNNs lies in their ability to learn expressive node representations, which are crucial for downstream tasks such as node classification, link prediction, and graph classification. For graph-level tasks, including graph classification, a readout or pooling operation is applied to aggregate node-level embeddings into a single graph representation:

\begin{equation*}
\boldsymbol{h}_G = \text{POOL} \left( \{ \boldsymbol{h}_v^{(T)} : v \in G \} \right),
\end{equation*}

where \( \boldsymbol{h}_v^{(T)} \) is the final representation of node \( v \) after \( T \) message passing iterations, and \texttt{POOL} denotes an aggregation function such as summation, mean, max, or a more sophisticated pooling scheme \cite{xu2018powerful}.

In this work, we focus on improving the quality of initial node features to be used within this message passing framework. By integrating structurally meaningful attributes—such as those derived from Network Control Theory—we aim to enhance the expressiveness and predictive power of GNNs in feature-scarce environments.


\section{Numerical Evaluation}
\label{sec:evaluation}



\begin{figure*}[!t]
    \centering
    \caption{Rank Encoding: Comparison of different bin sizes (k) on the Reddit (left) and GitHub Stargazers (right) datasets.}
    \vspace{1em}
    \begin{minipage}{0.49\linewidth}
        \centering
        \includegraphics[width=\linewidth]{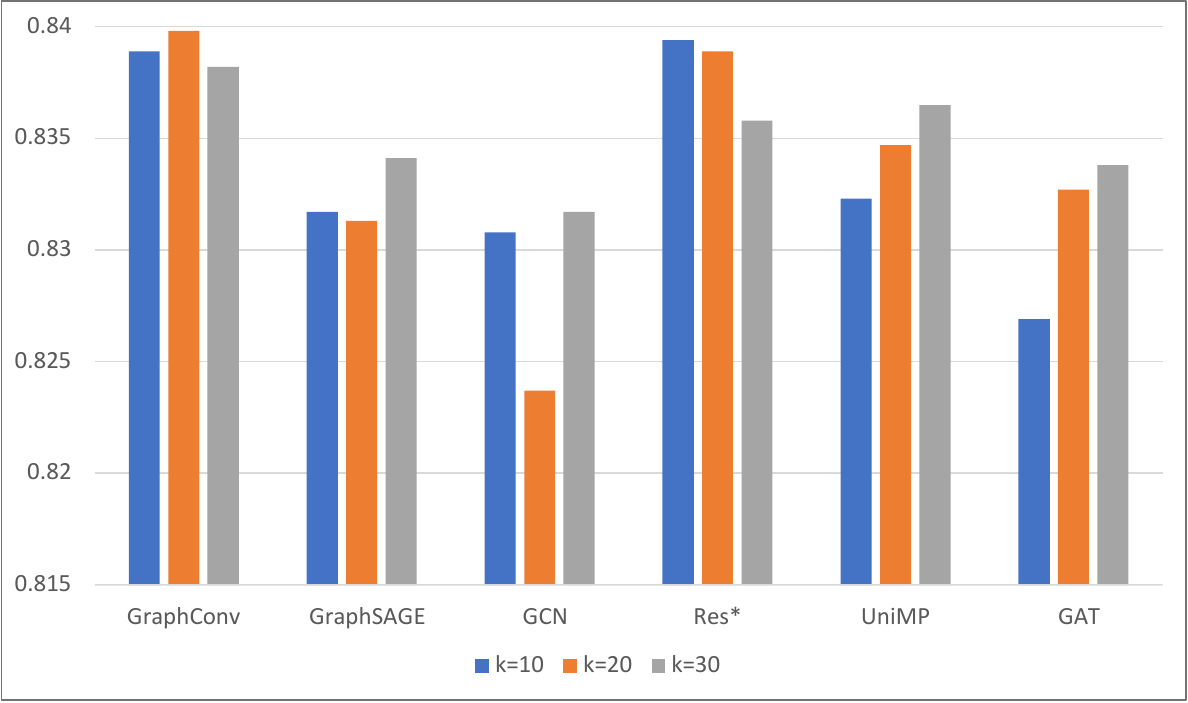}
        
        \label{fig:reddit}
    \end{minipage}
    \hfill
    \begin{minipage}{0.49\linewidth}
        \centering
        \includegraphics[width=\linewidth]{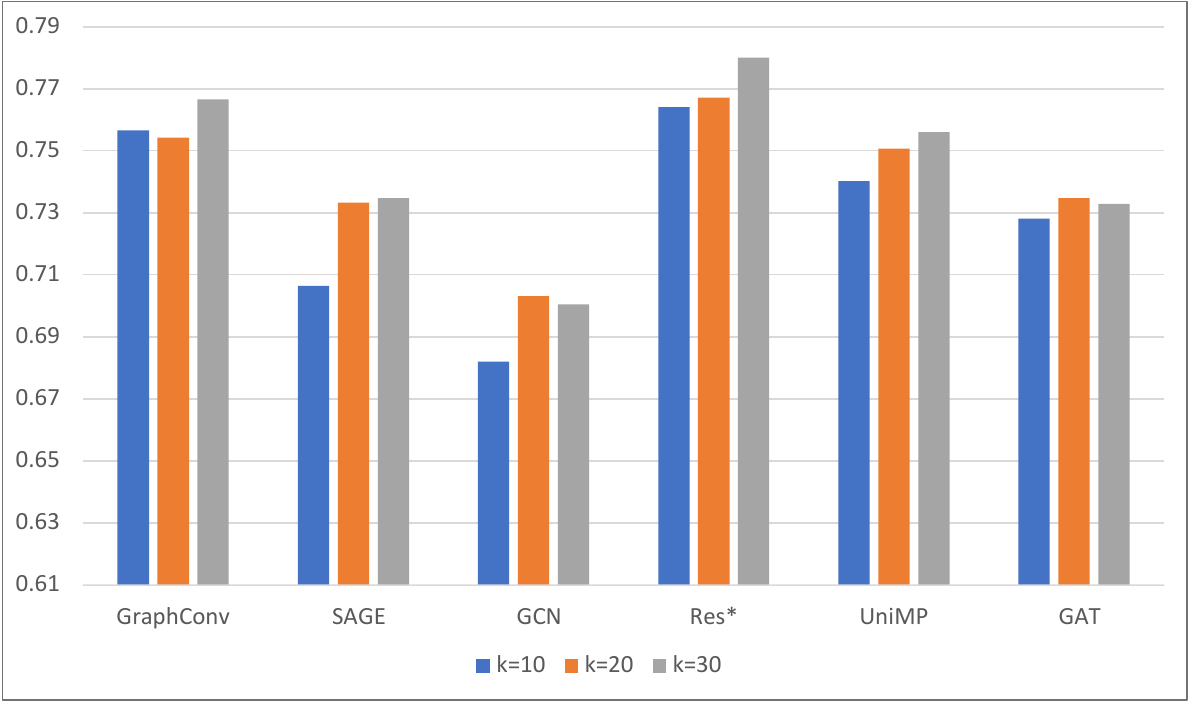}
        \label{fig:github_stargazers}
    \end{minipage}
    \label{fig:bins-comparison}
\end{figure*}

\begin{table*}[!htb]
\centering
\caption{ROC AUC Comparison (Mean $\pm$ Std) of Degree Encoding vs. Average Controllability \& Concatenated Rank Encoding. Bold indicates best results.}
\label{tab:results_deg_ac_concat}
\resizebox{\textwidth}{!}{
\begin{tabular}{|l|ccc|ccc|ccc|}
\hline
\multirow{2}{*}{Datasets} 
& \multicolumn{3}{c|}{\textbf{$k-$GNN}} 
& \multicolumn{3}{c|}{\textbf{GraphSAGE}} 
& \multicolumn{3}{c|}{\textbf{GCN}} \\
\cline{2-10}
& \textit{deg} & AC & Concat & \textit{deg} & AC & Concat & \textit{deg} & AC & Concat \\
\hline

Reddit Threads        
& 83.14 $\pm$ 0.11 & \textbf{83.89 $\pm$ 0.03} & 83.70 $\pm$ 0.08 
& 82.86 $\pm$ 0.24 & 83.17 $\pm$ 0.09 & \textbf{83.39 $\pm$ 0.06} 
& 82.74 $\pm$ 0.22 & 83.08 $\pm$ 0.30 & \textbf{83.19 $\pm$ 0.04}  \\ 

GitHub Stargazers     
& 71.47 $\pm$ 0.56 & \textbf{75.66 $\pm$ 0.51} & 74.88 $\pm$ 0.67 
& 68.70 $\pm$ 1.40 & 70.64 $\pm$ 0.41 & \textbf{74.48 $\pm$ 0.37} 
& 70.53 $\pm$ 0.52 & 68.20 $\pm$ 0.35 & \textbf{72.77 $\pm$ 0.50}  \\ 

Deezer Egos      
& 53.93 $\pm$ 1.26 & 52.61 $\pm$ 1.69 & \textbf{55.16 $\pm$ 0.66}
& \textbf{53.29 $\pm$ 1.20} & 51.27 $\pm$ 0.56 & 51.89 $\pm$ 0.63 
& 51.31 $\pm$ 2.20 & \textbf{53.21 $\pm$ 0.56} & 51.06 $\pm$ 0.74  \\ 

Twitch Egos      
& 65.26 $\pm$ 0.05 & 65.10 $\pm$ 0.11 & \textbf{65.26 $\pm$ 0.13} 
& 64.68 $\pm$ 0.05 & 64.80 $\pm$ 0.11 & \textbf{64.85 $\pm$ 0.15} 
& \textbf{64.71 $\pm$ 0.10} & 64.67 $\pm$ 0.16 & 64.61 $\pm$ 0.06  \\ 
\hline\hline 


\multirow{2}{*}{Datasets} 
& \multicolumn{3}{c|}{\textbf{UniMP}} 
& \multicolumn{3}{c|}{\textbf{ResGatedGCN}} 
& \multicolumn{3}{c|}{\textbf{GAT}} \\
\cline{2-10}
& \textit{deg} & AC & Concat & \textit{deg} & AC & Concat & \textit{deg} & AC & Concat \\
\hline 

Reddit Threads        
& \textbf{83.55 $\pm$ 0.05} & 83.23 $\pm$ 0.16 & 83.50 $\pm$ 0.06 
& 83.68 $\pm$ 0.08 & 83.94 $\pm$ 0.04 & \textbf{83.95 $\pm$ 0.04} 
& 83.44 $\pm$ 0.03 & 82.69 $\pm$ 0.22 & \textbf{83.45 $\pm$ 0.06} \\ 

GitHub Stargazers     
& 68.75 $\pm$ 0.65 & 74.02 $\pm$ 0.38 & \textbf{74.60 $\pm$ 0.37}
& 72.63 $\pm$ 0.53 & \textbf{76.41 $\pm$ 0.71} & 75.47 $\pm$ 0.41
& 67.30 $\pm$ 1.82 & \textbf{72.82 $\pm$ 0.46} & 72.50 $\pm$ 0.31 \\ 

Deezer Egos      
& \textbf{54.55 $\pm$ 0.88} & 51.60 $\pm$ 1.91 & 50.93 $\pm$ 1.09
& 54.30 $\pm$ 1.55 & 54.31 $\pm$ 1.18 & \textbf{54.53 $\pm$ 0.73}
& \textbf{53.30 $\pm$ 1.57} & 51.57 $\pm$ 1.80 & 51.96 $\pm$ 1.08 \\ 

Twitch Egos      
& 64.76 $\pm$ 0.09 & \textbf{65.03 $\pm$ 0.12} & 64.82 $\pm$ 0.10
& 65.28 $\pm$ 0.12 & \textbf{65.32 $\pm$ 0.08} & 65.27 $\pm$ 0.24
& \textbf{65.16 $\pm$ 0.05} & 65.09 $\pm$ 0.09 & 64.97 $\pm$ 0.04 \\ 
\hline
\end{tabular}%
}
\end{table*}

In this section, we evaluate the effectiveness of our proposed methods through two distinct experimental settings. First, we conduct a comparative analysis between a feature set that includes average controllability, closeness centrality, betweenness centrality, and eigenvector centrality, and the baseline one-hot-degree encoding method (see Table \ref{tab:results_deg_nct}). This comparison aims to evaluate the ability of different centrality metrics to capture the topological information of the graph within the feature space. In the subsequent experiments, we examine the performance of the proposed rank encoding method in two configurations: encoding only average controllability and concatenating the encoding of multiple centrality metrics (see Table \ref{tab:results_deg_ac_concat}). The following sections provide a detailed discussion of the experimental setup and results.

\subsection{Datasets}

We consider the following four widely used social network benchmark datasets in our experimental setup.

\textbf{Reddit Threads:} This dataset comprises an assortment of threads extracted from the Reddit platform, all of which were gathered during the month of May 2018. It encompasses both discussion and non-discussion based threads, presenting a diverse range of community interactions. The task is to distinguish between threads that facilitate discussion and those that do not, thereby classifying the content based on its conversational nature and potential for user engagement \cite{karateclub}.

\textbf{GitHub Stargazers:} This is a social network dataset of developers who have starred notable machine learning and web development repositories on GitHub. The task involves classifying these networks to ascertain if they correspond to web development or machine learning repositories based on their stargazing activities \cite{karateclub}. 

\textbf{Twitch Egos:} This dataset consists of ego-networks of Twitch users who participated in the platform's partnership program in April 2018. Nodes represent users, while edges denote friendships. The classification task aims to predict whether an ego user plays a single game or multiple games based on their ego-network structure. Users who play a single game typically exhibit denser ego-net connections \cite{karateclub}.

\textbf{Deezer Egos:} This dataset contains ego-networks of Eastern European users collected from the music streaming platform Deezer in February 2020. Nodes correspond to users, and edges represent mutual follower relationships. The classification task involves predicting the gender of the ego user based on the structural properties of their ego network \cite{karateclub}. The dataset statistics are presented in Table \ref{tab:dataset-stats}.

\subsection{Experimental Setup}

We consider six well-known baselines graph convolution methods that include $k-$GNN (GraphConv) \cite{morris2019weisfeiler}, GraphSAGE \cite{hamilton2017inductive}, GCN \cite{kipf2016semi}, Transformer Convolution (UniMP) \cite{shi2020masked}, Residual Gated Graph Convolution (ResGatedGCN) \cite{bresson2017residual} and Graph Attention Network (GAT) \cite{velivckovic2017graph}.

The proposed learning framework comprises three layers of GNNs, each containing $64$ hidden units. Post these layers, Sort Aggregation \cite{zhang2018end} is applied. Following the aggregation step, we employ two layers of 1D convolution complemented by Max Pooling. This is succeeded by a multi-layer perceptron with two layers, each having 32 hidden neurons. For model evaluation, we resort to 10-fold cross-validation, training each model for $100$ epochs. We set the learning rate to $1e^{-4}$ and the weight decay to $5e^{-2}$. All experiments are conducted on a Lambda machine equipped with an AMD Ryzen Threadripper PRO $5995WX 64-$Core CPU, 512 GB RAM, and an NVIDIA RTX 6000 GPU with 48 GB of memory.

For rank encoding, the input parameters include the number of bins $k$ and the range of the histogram. In our experimental setup, we evaluate different values of $k$, specifically $10, 20$, and $30$. The results presented in Tables \ref{tab:results_deg_nct} and \ref{tab:results_deg_ac_concat} correspond to $k = 10$. In the concatenation experiment, additional centrality metrics were incorporated, each encoded with $k = 10$, and subsequently concatenated to form a feature vector of size 50 for each node. The histogram range was consistently defined by the minimum and maximum values throughout all experiments. 
\begin{figure*}[!t]
    \centering
    \caption{Rank Encoding: Comparison of different bin sizes on the Deezer Egonet (left) and Twitch (right) datasets.}
    \begin{minipage}{0.49\linewidth}
        \centering
        \includegraphics[width=\linewidth]{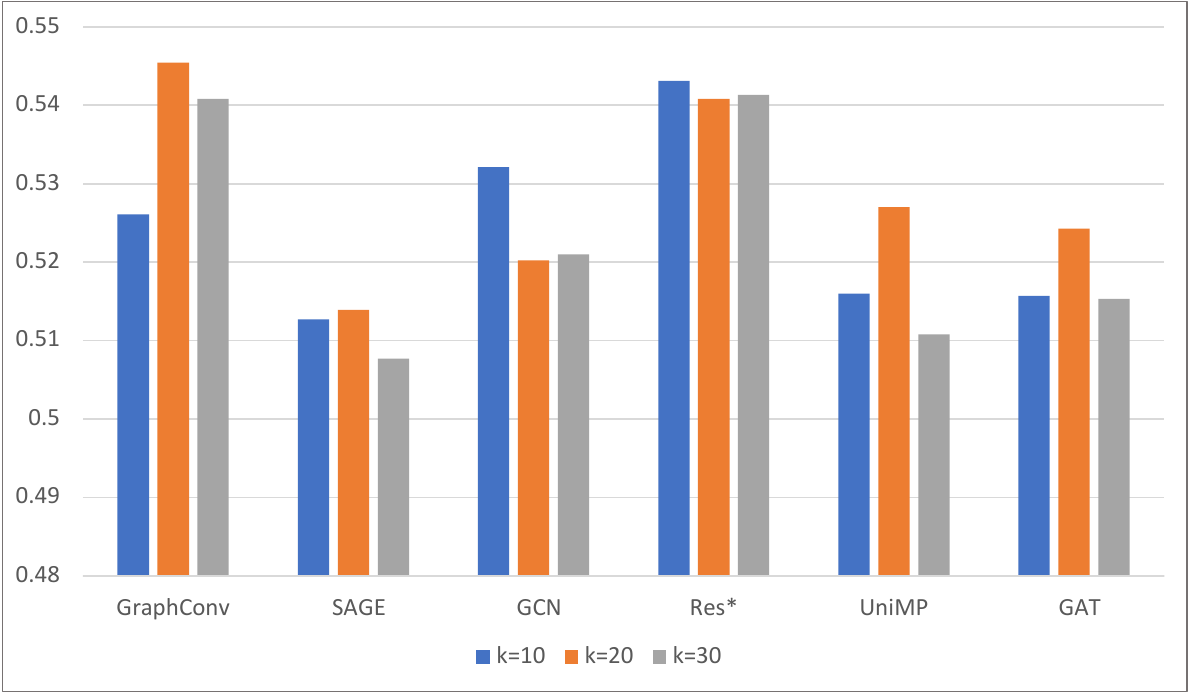}
        
        \label{fig:deezer}
    \end{minipage}
    \hfill
    \begin{minipage}{0.49\linewidth}
        \centering
        \includegraphics[width=\linewidth]{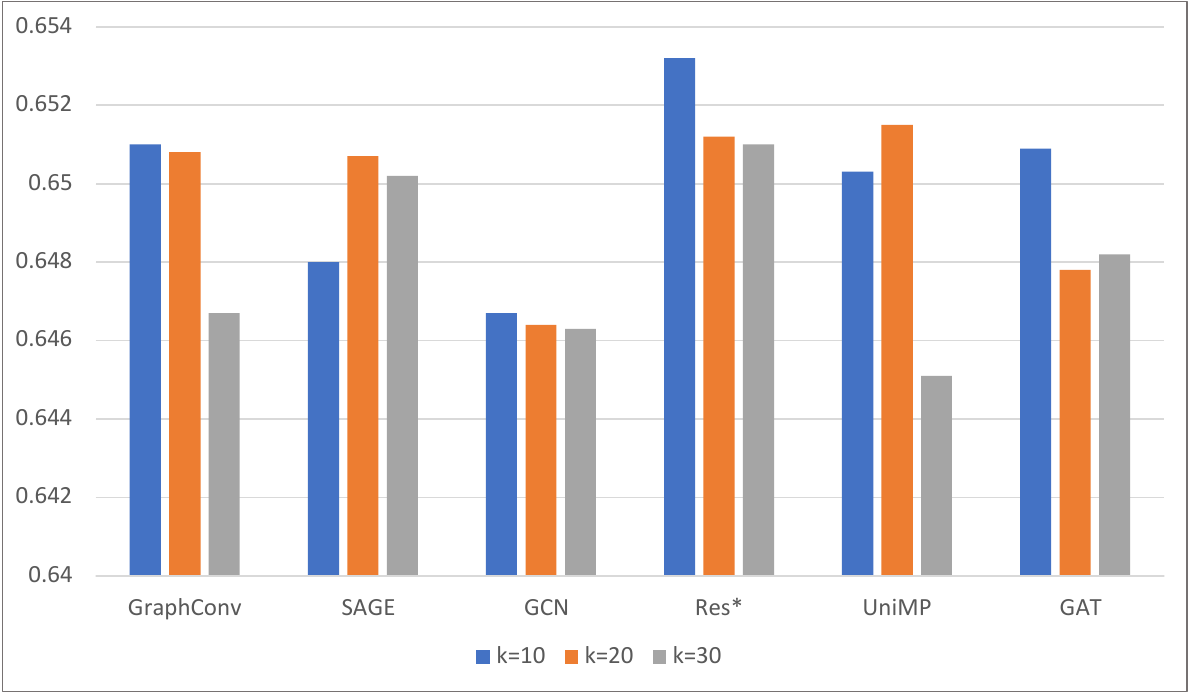}
    \end{minipage}
    
    \label{fig:bins-comparison-twitch}
\end{figure*}


\textbf{Evaluation Metric (ROC AUC):} The Receiver Operating Characteristic Area Under the Curve (ROC AUC) is a widely used performance metric for classification tasks, particularly when dealing with imbalanced datasets. It measures the model’s ability to distinguish between the positive and negative classes across all possible classification thresholds. The ROC curve plots the True Positive Rate (sensitivity) against the False Positive Rate (1 – specificity), and the area under this curve (AUC) provides a single scalar value summarizing the model's discriminative power. An AUC of 0.5 indicates no better than random guessing, while an AUC of 1.0 represents perfect classification.

\subsection{Results with Full Set of Features}
We present the ROC AUC (Receiver Operating Characteristic Area Under the Curve) classification results for all datasets using the full set of features in Table \ref{tab:results_deg_nct}. We maintained consistent architecture and experimental settings for evaluation, comparing the performance of both one-hot-degree encoding (deg) and the full set of feature denoted as NCT-EFA.
Overall, NCT-EFA demonstrates superior performance across most datasets and architectures, with particularly notable improvements observed in the GitHub Stargazers dataset. Specifically, NCT-EFA achieves a substantial $6.54\%$ increase in ROC AUC with $k-$GNN, a $7.66\%$ improvement with GraphSAGE, and a $5.53\%$ gain with ResGatedGCN. For Reddit Threads, while the performance gains are relatively moderate, NCT-EFA consistently outperforms degree encoding, with improvements ranging from $0.15\%$ to $0.90\%$ across different architectures. This suggests that although the dataset’s structural properties already contribute significantly to classification, NCT-EFA enhances feature representation without inducing drastic changes. In contrast, results for Deezer Egos dataset show that degree encoding outperforms NCT-EFA in most cases, with declines of up to $3.47\%$ in ROC AUC. This may be attributed to the nature of gender prediction in social networks, where degree-based information plays a more crucial role than the feature transformation provided by NCT-EFA. Lastly, for Twitch Egos, NCT-EFA consistently outperforms degree encoding across all architectures, with improvements ranging from $0.11\%$ to $0.59\%$. This suggests that the proposed method is effective in capturing user interaction patterns for classifying gaming behavior.

These results underscore the effectiveness of NCT-EFA in enhancing classification performance across a different datasets. Notably, NCT-EFA performs exceptionally well on the GitHub Stargazers dataset, likely due to the importance of node features in distinguishing between machine learning and web development repositories. The distinct node features of users engaged with these types of repositories make classification more reliant on feature-based information. Meanwhile, classification tasks might rely more heavily on graph topology for other datasets.


\subsection{Numerical Results with Rank Encoding}

To assess the effectiveness of rank encoding compared to degree encoding, we conducted additional experiments using two variations of rank encoding: average controllability rank encoding (AC) and combined rank encoding with multiple centrality measures (Concat), including average controllability), as discussed in Section \ref{subsec:node-features-construction}. We maintained the same architecture and experimental setup, replacing degree-based node features with these alternative encoding to evaluate their impact on classification performance. The results are summarized in Table~\ref{tab:results_deg_ac_concat}.

Our results indicate that both AC and Concat encoding consistently outperform degree encoding in datasets where node features play a crucial role in classification. For GitHub Stargazers, AC alone yields significant gains, with $4.19\%$ improvement in ROC AUC for GCN, $5.78\%$ for GraphSAGE, and $5.94\%$ for UniMP. Furthermore, Concat encoding further enhances performance, achieving $5.66\%$, $6.21\%$, and $6.83\%$ improvements, respectively. This suggests that while average controllability alone provides valuable structural information, incorporating additional centrality measures refines the feature representation, leading to stronger classification performance. For Reddit Threads, the improvements are relatively moderate but still consistent. AC encoding improves classification by $0.15\%$ to $0.90\%$, while Concat encoding provides slightly higher gains, reinforcing the idea that structural rankings enhance feature representation without drastically altering the dataset’s inherent classification properties.


In contrast, for Deezer Egos, degree encoding remains the superior approach, with AC and Concat encoding leading to performance declines of up to \$3.47\%\$ in ROC AUC. Notably, however, the overall performance across all encoding strategies remains close to random, with ROC AUC values ranging from approximately 51\% to 55\%. This indicates that gender prediction from this particular social network is inherently challenging, likely due to the weak or noisy signal associated with gender in the structural patterns of ego networks. In other words, the network topology in Deezer Egos may not meaningfully reflect gender-specific connectivity, limiting the potential of the models to extract relevant features for this task. The modest performance gains observed with degree encoding suggest that absolute degree may capture weak demographic correlations—such as differences in user activity levels or connection tendencies—but the overall low discriminative power across methods points to an underlying lack of informative structure in the data for the gender prediction task.

Overall, these results indicate that encoding average controllability significantly enhances the performance of all the GNN models. While AC alone is effective, Concat encoding provides the most robust performance gains across datasets, confirming that leveraging a broader structural perspective enhances classification outcomes. However, degree encoding remains more effective in social networks where connectivity is a dominant predictive factor.

In the rank encoding experiment, we also compare different bin sizes $(k=10, 20$, and $30$ to evaluate the impact of feature dimensionality on model performance. The results, presented in Figures \(\ref{fig:bins-comparison}\) and \(\ref{fig:bins-comparison-twitch}\), demonstrate that model performance varies significantly across different values of \( k \) and datasets, with no single configuration consistently outperforming the others. However, for the Reddit dataset, \( k = 30 \) achieves the best results in \( 5 \) out of \( 6 \) cases, while for the GitHub Stargazers dataset, it performs best in \( 4 \) out of \( 6 \) cases. For the remaining two datasets, the differences in performance across different values of \( k \) are minimal, suggesting that the choice of \( k \) has no substantial impact on the overall model performance.



\section{Conclusions and Future Work}
\label{sec:conclusion}

In this work, we introduced a novel feature initialization approach based on average controllability to enhance the performance of GNNs in graph classification tasks. Our method incorporated average controllability, closeness centrality, betweenness centrality, and eigenvector centrality as node features, evaluated across six different GNN architectures. Furthermore, we proposed a rank encoding scheme that transforms average controllability—or any other graph metric—into expressive, fixed-dimensional node representations, further strengthening the effectiveness of our approach. Experimental results across multiple GNN variants and diverse social network datasets demonstrated consistent performance improvements, underscoring the efficacy of the proposed methods.

These findings open several promising directions for future research. One potential avenue is the integration of multiple controllability metrics to further enrich node features and improve GNN performance. Additionally, embedding controllability information directly into the message-passing mechanisms of GNNs presents an intriguing research direction. Another compelling extension is to leverage controllability metrics for constructing expressive graph-level embeddings, potentially advancing the quality of graph representation learning.

The integration of controllability metrics with GNNs holds significant promise for advancing graph machine learning. We believe this work offers a foundational step toward bridging these domains, paving the way for more expressive and principled graph learning methodologies.

\textbf{Computational complexity:}

The computation of average controllability involves numerical integration of the finite-horizon controllability Gramian, which requires repeated matrix multiplications and exponentiations. While this process avoids solving the Lyapunov equation directly, it still incurs a worst-case time complexity of \( O(n^3) \), where \( n \) is the number of nodes in the graph. In practice, this computation remains tractable for the graph sizes considered in our study. For instance, calculating the average controllability for a random Erdős–Rényi graph with 100 nodes and approximately 50\% edge density takes around 1 second on our machine. In the Reddit dataset, although there are over 200,000 ego-graphs, each graph is small (with a maximum of 97 nodes). By parallelizing the computation across 128 CPU cores, we were able to process the entire Reddit dataset in under an hour. These benchmarks demonstrate the scalability of our approach for medium-scale graphs and practical datasets. Further acceleration through low-rank approximations, incremental updates, or spectral methods is a promising direction for future work, particularly for large-scale or dynamic graph scenarios.


\section{Acknolwdgements}
This work is supported by the National Science Foundation under Grant Numbers 2325416 and 2325417.
\bibliographystyle{IEEEtran}
\bibliography{references}

\begin{thebibliography}{10}
\providecommand{\url}[1]{#1}
\csname url@samestyle\endcsname
\providecommand{\newblock}{\relax}
\providecommand{\bibinfo}[2]{#2}
\providecommand{\BIBentrySTDinterwordspacing}{\spaceskip=0pt\relax}
\providecommand{\BIBentryALTinterwordstretchfactor}{4}
\providecommand{\BIBentryALTinterwordspacing}{\spaceskip=\fontdimen2\font plus
\BIBentryALTinterwordstretchfactor\fontdimen3\font minus \fontdimen4\font\relax}
\providecommand{\BIBforeignlanguage}[2]{{%
\expandafter\ifx\csname l@#1\endcsname\relax
\typeout{** WARNING: IEEEtran.bst: No hyphenation pattern has been}%
\typeout{** loaded for the language `#1'. Using the pattern for}%
\typeout{** the default language instead.}%
\else
\language=\csname l@#1\endcsname
\fi
#2}}
\providecommand{\BIBdecl}{\relax}
\BIBdecl

\bibitem{gu2015controllability}
S.~Gu, F.~Pasqualetti, M.~Cieslak, Q.~K. Telesford, A.~B. Yu, A.~E. Kahn, J.~D. Medaglia, J.~M. Vettel, M.~B. Miller, S.~T. Grafton \emph{et~al.}, ``Controllability of structural brain networks,'' \emph{Nature communications}, vol.~6, no.~1, p. 8414, 2015.

\bibitem{friedland2012control}
B.~Friedland, \emph{Control system design: an introduction to state-space methods}.\hskip 1em plus 0.5em minus 0.4em\relax Courier Corporation, 2012.

\bibitem{pasqualetti2014controllability}
F.~Pasqualetti, S.~Zampieri, and F.~Bullo, ``Controllability metrics, limitations and algorithms for complex networks,'' \emph{IEEE Transactions on Control of Network Systems}, vol.~1, no.~1, pp. 40--52, 2014.

\bibitem{hamilton2020graph}
W.~L. Hamilton, \emph{Graph representation learning}.\hskip 1em plus 0.5em minus 0.4em\relax Morgan \& Claypool Publishers, 2020.

\bibitem{kipf2016semi}
T.~N. Kipf and M.~Welling, ``Semi-supervised classification with graph convolutional networks,'' in \emph{International Conference on Learning Representations}, 2017.

\bibitem{hamilton2017inductive}
W.~Hamilton, Z.~Ying, and J.~Leskovec, ``Inductive representation learning on large graphs,'' \emph{Advances in neural information processing systems}, vol.~30, 2017.

\bibitem{xu2018powerful}
K.~Xu, W.~Hu, J.~Leskovec, and S.~Jegelka, ``How powerful are graph neural networks?'' in \emph{International Conference on Learning Representations}, 2019.

\bibitem{kalman1960contributions}
R.~E. Kalman \emph{et~al.}, ``Contributions to the theory of optimal control,'' \emph{Bol. soc. mat. mexicana}, vol.~5, no.~2, pp. 102--119, 1960.

\bibitem{gasteiger2019diffusion}
J.~Gasteiger, S.~Wei{\ss}enberger, and S.~G{\"u}nnemann, ``Diffusion improves graph learning,'' \emph{Advances in neural information processing systems}, vol.~32, 2019.

\bibitem{freeman2002centrality}
L.~C. Freeman \emph{et~al.}, ``Centrality in social networks: Conceptual clarification,'' \emph{Social network: critical concepts in sociology. Londres: Routledge}, vol.~1, pp. 238--263, 2002.

\bibitem{parkes2021network}
L.~Parkes, T.~M. Moore, M.~E. Calkins, M.~Cieslak, D.~R. Roalf, D.~H. Wolf, R.~C. Gur, R.~E. Gur, T.~D. Satterthwaite, and D.~S. Bassett, ``Network controllability in transmodal cortex predicts positive psychosis spectrum symptoms,'' \emph{Biological Psychiatry}, vol.~90, no.~6, pp. 409--418, 2021.

\bibitem{said2024improving}
A.~Said, O.~U. Ahmad, W.~Abbas, M.~Shabbir, and X.~Koutsoukos, ``Improving graph machine learning performance through feature augmentation based on network control theory,'' in \emph{2024 32nd Mediterranean Conference on Control and Automation (MED)}.\hskip 1em plus 0.5em minus 0.4em\relax IEEE, 2024, pp. 322--327.

\bibitem{said2023network}
------, ``Network controllability perspectives on graph representation,'' \emph{IEEE Transactions on Knowledge and Data Engineering}, vol.~36, no.~8, pp. 4116--4128, 2023.

\bibitem{parkes2023using}
L.~Parkes, J.~Z. Kim, J.~Stiso, J.~K. Brynildsen, M.~Cieslak, S.~Covitz, R.~E. Gur, R.~C. Gur, F.~Pasqualetti, R.~T. Shinohara \emph{et~al.}, ``Using network control theory to study the dynamics of the structural connectome,'' \emph{bioRxiv}, 2023.

\bibitem{kim2020linear}
J.~Z. Kim and D.~S. Bassett, ``Linear dynamics and control of brain networks,'' \emph{Neural Engineering}, pp. 497--518, 2020.

\bibitem{yang2024control}
J.~Yang, R.~Ding, F.~Ji, H.~Wang, and L.~Xie, ``Control the gnn: Utilizing neural controller with lyapunov stability for test-time feature reconstruction,'' \emph{arXiv preprint arXiv:2410.09708}, 2024.

\bibitem{kriege2020survey}
N.~M. Kriege, F.~D. Johansson, and C.~Morris, ``A survey on graph kernels,'' \emph{Applied Network Science}, vol.~5, no.~1, pp. 1--42, 2020.

\bibitem{togninalli2019wasserstein}
M.~Togninalli, E.~Ghisu, F.~Llinares-L{\'o}pez, B.~Rieck, and K.~Borgwardt, ``Wasserstein weisfeiler-lehman graph kernels,'' \emph{Advances in neural information processing systems}, vol.~32, 2019.

\bibitem{kondor2016multiscale}
R.~Kondor and H.~Pan, ``The multiscale laplacian graph kernel,'' \emph{Advances in neural information processing systems}, vol.~29, 2016.

\bibitem{defferrard2016convolutional}
M.~Defferrard, X.~Bresson, and P.~Vandergheynst, ``Convolutional neural networks on graphs with fast localized spectral filtering,'' \emph{Advances in neural information processing systems}, vol.~29, 2016.

\bibitem{bruna2013spectral}
J.~Bruna, W.~Zaremba, A.~Szlam, and Y.~Lecun, ``Spectral networks and locally connected networks on graphs,'' in \emph{International Conference on Learning Representations (ICLR2014)}, 2014, pp. http--openreview.

\bibitem{bresson2017residual}
X.~Bresson and T.~Laurent, ``Residual gated graph convnets,'' \emph{arXiv preprint arXiv:1711.07553}, 2017.

\bibitem{velivckovic2017graph}
P.~Veli{\v{c}}kovi{\'c}, G.~Cucurull, A.~Casanova, A.~Romero, P.~Li{\`o}, and Y.~Bengio, ``Graph attention networks,'' in \emph{International Conference on Learning Representations}, 2018.

\bibitem{morris2019weisfeiler}
C.~Morris, M.~Ritzert, M.~Fey, W.~L. Hamilton, J.~E. Lenssen, G.~Rattan, and M.~Grohe, ``Weisfeiler and leman go neural: Higher-order graph neural networks,'' in \emph{Proceedings of the AAAI conference on artificial intelligence}, vol.~33, no.~01, 2019, pp. 4602--4609.

\bibitem{yun2019graph}
S.~Yun, M.~Jeong, R.~Kim, J.~Kang, and H.~J. Kim, ``Graph transformer networks,'' \emph{Advances in neural information processing systems}, vol.~32, 2019.

\bibitem{ying2018hierarchical}
Z.~Ying, J.~You, C.~Morris, X.~Ren, W.~Hamilton, and J.~Leskovec, ``Hierarchical graph representation learning with differentiable pooling,'' \emph{Advances in neural information processing systems}, vol.~31, 2018.

\bibitem{diehl2019edge}
F.~Diehl, ``Edge contraction pooling for graph neural networks,'' \emph{CoRR}, 2019.

\bibitem{wu2020comprehensive}
Z.~Wu, S.~Pan, F.~Chen, G.~Long, C.~Zhang, and S.~Y. Philip, ``A comprehensive survey on graph neural networks,'' \emph{IEEE transactions on neural networks and learning systems}, no.~1, 2020.

\bibitem{cai2018simple}
C.~Cai and Y.~Wang, ``A simple yet effective baseline for non-attributed graph classification,'' \emph{arXiv preprint arXiv:1811.03508}, 2018.

\bibitem{lee2019attention}
J.~B. Lee, R.~A. Rossi, S.~Kim, N.~K. Ahmed, and E.~Koh, ``Attention models in graphs: A survey,'' \emph{ACM Transactions on Knowledge Discovery from Data (TKDD)}, vol.~13, no.~6, pp. 1--25, 2019.

\bibitem{belkin2003laplacian}
M.~Belkin and P.~Niyogi, ``Laplacian eigenmaps for dimensionality reduction and data representation,'' \emph{Neural computation}, vol.~15, no.~6, pp. 1373--1396, 2003.

\bibitem{dwivedi2023benchmarking}
V.~P. Dwivedi, C.~K. Joshi, A.~T. Luu, T.~Laurent, Y.~Bengio, and X.~Bresson, ``Benchmarking graph neural networks,'' \emph{Journal of Machine Learning Research}, vol.~24, no.~43, pp. 1--48, 2023.

\bibitem{ribeiro2017struc2vec}
L.~F. Ribeiro, P.~H. Saverese, and D.~R. Figueiredo, ``struc2vec: Learning node representations from structural identity,'' in \emph{Proceedings of the 23rd ACM SIGKDD international conference on knowledge discovery and data mining}, 2017, pp. 385--394.

\bibitem{ahmed2019role2vec}
N.~K. Ahmed, R.~A. Rossi, J.~B. Lee, T.~L. Willke, R.~Zhou, X.~Kong, and H.~Eldardiry, ``role2vec: Role-based network embeddings,'' \emph{Proc. DLG KDD}, pp. 1--7, 2019.

\bibitem{donnat2018learning}
C.~Donnat, M.~Zitnik, D.~Hallac, and J.~Leskovec, ``Learning structural node embeddings via diffusion wavelets,'' in \emph{Proceedings of the 24th ACM SIGKDD international conference on knowledge discovery \& data mining}, 2018, pp. 1320--1329.

\bibitem{ma2021unified}
Y.~Ma, X.~Liu, T.~Zhao, Y.~Liu, J.~Tang, and N.~Shah, ``A unified view on graph neural networks as graph signal denoising,'' in \emph{Proceedings of the 30th ACM international conference on information \& knowledge management}, 2021, pp. 1202--1211.

\bibitem{ahmad2024control}
O.~U. Ahmad, A.~Said, M.~Shabbir, X.~Koutsoukos, and W.~Abbas, ``Control-based graph embeddings with data augmentation for contrastive learning,'' in \emph{2024 American Control Conference (ACC)}.\hskip 1em plus 0.5em minus 0.4em\relax IEEE, 2024, pp. 27--32.

\bibitem{wasserman1994social}
S.~Wasserman, ``Social network analysis: Methods and applications,'' \emph{The Press Syndicate of the University of Cambridge}, 1994.

\bibitem{karateclub}
B.~Rozemberczki, O.~Kiss, and R.~Sarkar, ``{Karate Club: An API Oriented Open-source Python Framework for Unsupervised Learning on Graphs},'' in \emph{Proceedings of the 29th ACM International Conference on Information and Knowledge Management (CIKM '20)}.\hskip 1em plus 0.5em minus 0.4em\relax ACM, 2020, p. 3125–3132.

\bibitem{shi2020masked}
Y.~Shi, Z.~Huang, S.~Feng, H.~Zhong, W.~Wang, and Y.~Sun, ``Masked label prediction: Unified message passing model for semi-supervised classification,'' in \emph{Proceedings of the Thirtieth International Joint Conference on Artificial Intelligence}.\hskip 1em plus 0.5em minus 0.4em\relax International Joint Conferences on Artificial Intelligence Organization, 2021, pp. 1548--1554.

\bibitem{zhang2018end}
M.~Zhang, Z.~Cui, M.~Neumann, and Y.~Chen, ``An end-to-end deep learning architecture for graph classification,'' in \emph{Proceedings of the AAAI conference on artificial intelligence}, vol.~32, no.~1, 2018.

\end{thebibliography}

\end{document}